\documentclass[11pt]{article}

\usepackage[preprint]{acl}

\usepackage{times}
\usepackage{tikz}
\usepackage{contour}
\usepackage{latexsym}
\usepackage{amsmath} 
\usepackage[most]{tcolorbox}
\usepackage{amsfonts}
\usepackage{xcolor}
\usepackage{soul} %
\usepackage{fancybox}

\usepackage[T1]{fontenc}

\usepackage[utf8]{inputenc}

\usepackage{microtype}
\usepackage{adjustbox}

\usepackage{inconsolata}

\usepackage{graphicx}
\usepackage{booktabs}

\usepackage{enumitem}

\newtcbox{\myhighlight}{on line,
  colframe=orange, colback=white,
  boxrule=0.5pt, boxsep=1pt, arc=2pt,
  left=2pt, right=2pt, top=1pt, bottom=1pt,
  enhanced}

\newtcolorbox{wraphighlight}{%
  enhanced,
  breakable,
  colback=white,       %
  colframe=black,     %
  boxrule=0.6pt,       %
  arc=2pt,             %
  left=4pt, right=4pt, top=2pt, bottom=2pt
}

\newcommand{\hlblue}[1]{\sethlcolor{blue!10}\hl{#1}}
\newcommand{\hlgreen}[1]{\sethlcolor{green!10}\hl{#1}}
\newcommand{\hlyellow}[1]{\sethlcolor{yellow!15}\hl{#1}}
\newcommand{\hlred}[1]{\sethlcolor{red!10}\hl{#1}}
\newcommand{\hlcyan}[1]{\sethlcolor{cyan!10}\hl{#1}}
\newcommand{\hlmagenta}[1]{\sethlcolor{magenta!10}\hl{#1}}
\newcommand{\hlorange}[1]{\sethlcolor{orange!10}\hl{#1}}

\title{Correct, Concise and Complete:\\Multi-stage Training For Adaptive Reasoning}

\author{
  Nathanaël Carraz Rakotonirina\thanks{Work conducted during an internship at AWS AI Labs.}~$^\diamondsuit$ \quad  Ren Pang~$^\clubsuit$ \quad  Neha Anna John~$^\clubsuit$ \\
  \textbf{Michael Bohlke-Schneider~$^\clubsuit$ \quad  Momchil Hardalov~$^\spadesuit$}  \\
  $^\diamondsuit$Universitat Pompeu Fabra \quad $^\clubsuit$AWS AI Labs \quad $^\spadesuit$Amazon AGI \\
    \texttt{nathanael.rakotonirina@upf.edu}\\
  \texttt{\{renpang, nehajohn, bohlkem, momchilh\}@amazon.com}}

\begin{document}
\maketitle
\begin{abstract}

The reasoning capabilities of large language models (LLMs) have improved substantially through increased test-time computation, typically in the form of intermediate tokens known as chain-of-thought (CoT). However, CoT often becomes unnecessarily long, increasing computation cost without actual accuracy gains or sometimes even degrading performance, a phenomenon known as ``\emph{overthinking}''. We propose a multi-stage efficient reasoning method that combines supervised fine-tuning---via rejection sampling or reasoning trace reformatting---with reinforcement learning using an adaptive length penalty. We introduce a lightweight reward function that penalizes tokens generated after the first correct answer but encouraging self-verification only when beneficial. We conduct a holistic evaluation across seven diverse reasoning tasks, analyzing the accuracy--response length trade-off. Our approach reduces response length by an average of 28\% for 8B models and 40\% for 32B models, while incurring only minor performance drops of 1.6 and 2.5 points, respectively. Despite its conceptual simplicity, it achieves a superior trade-off compared to more complex state-of-the-art efficient reasoning methods, scoring 76.6, in terms of the area under the Overthinking-Adjusted Accuracy curve ($\text{AUC}_{\text{OAA}}$)---5 points above the base model and 2.5 points above the second-best approach.

\end{abstract}

\section{Introduction}

Large language models (LLMs) achieve stronger performance on reasoning-intensive tasks, such as math and code generation, by increasing test-time computation~\citep{snell2025scaling, jaech2024openai, guo2025deepseek, openai2025}. Accuracy often improves as the model generates longer chains of thought (CoT). However, reasoning traces can also become unnecessarily long and often repetitive, yielding no additional gains, and in some cases even reducing accuracy, a phenomenon known as ``\emph{overthinking}'' \citep{chen2025do, wu2025more, yang2025towards}.

\begin{figure}[!t]
    \centering
    \includegraphics[width=1\columnwidth]{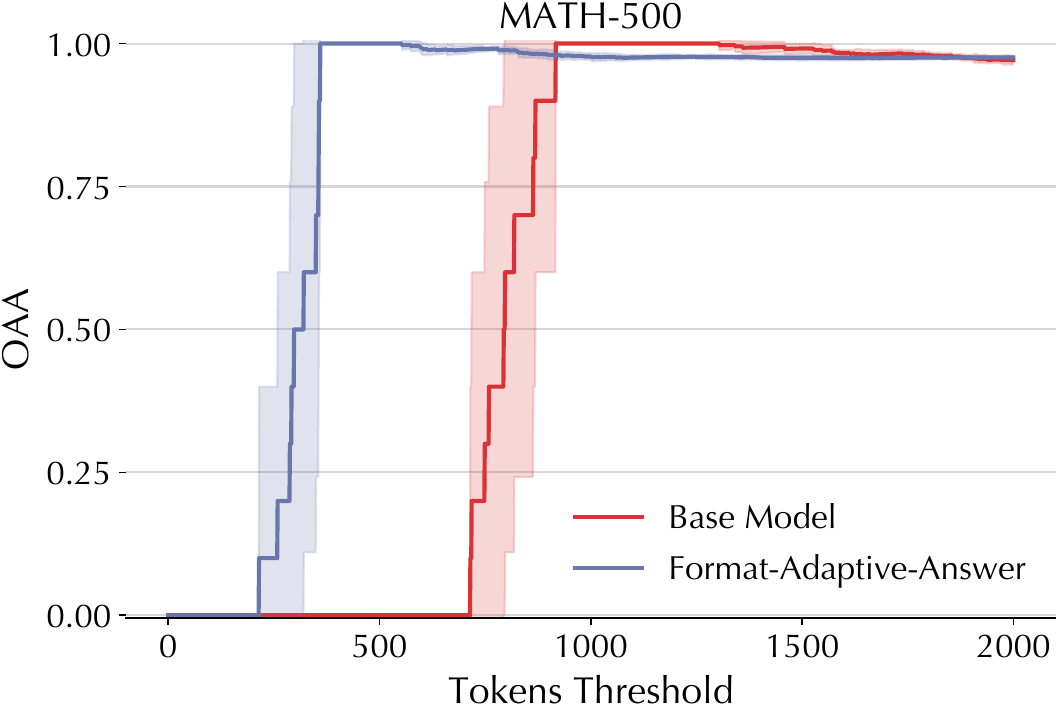}
    \caption{Overthinking-Adjusted Accuracy (OAA)~\citep{aggarwal2025optimalthinkingbench} as a function of the response length threshold on MATH-500 for Qwen3-8B. Our approach achieves similar accuracy with fewer tokens, leading to a larger area under the curve.}
    \label{fig:auc}
\end{figure}

To mitigate this, existing methods often impose a {predefined} thinking budget, {truncating the reasoning trace once the budget is reached}~\citep{yang2025towards} or {enforcing it as a hard constraint during reinforcement learning (RL) training}~\citep{hou2025thinkprune}. However, such {non-adaptive methods} are unable to {optimally balance accuracy and efficiency with respect to the response length}~\citep{snell2025scaling, wu2025more, yang2025towards}.

We introduce a multi-stage efficient reasoning framework that adaptively shortens response length while maintaining the base models' accuracy. Our method consists of supervised fine-tuning (SFT), followed by reinforcement learning with a length penalty. We construct the training dataset for the SFT stage using two approaches: rejection sampling, selecting the shortest correct response for each problem, and reformatting reasoning traces to omit additional summaries and provide the final answer. For the RL stage, we design a reward function that penalizes tokens generated after the first correct answer in the trace, encouraging concise yet complete reasoning traces that lead to the correct answer. It also incentivizes the model to perform self-verification only when necessary, as we show in our analysis.

We evaluate our methods on models of different sizes from the Qwen3 and DeepSeek families, using a wide range of reasoning benchmarks, including mathematics, science, code generation, question answering, and long-context tasks. Our approach significantly reduces response length while maintaining high accuracy. Furthermore, when measured using the area under the Overthinking-Adjusted Accuracy curve (OAA;~\citet{aggarwal2025optimalthinkingbench}), a unified metric that accounts for overthinking (see Figure~\ref{fig:auc}), our methods consistently improve both over the base models and state-of-the-art efficient reasoning approaches. 
Our main contributions are as follows:
\begin{itemize}[leftmargin=*, nosep]
    \item We propose a multi-stage efficient reasoning framework combining SFT via rejection sampling or via trace reformatting and RL with a length-penalizing reward that penalizes tokens generated after the first correct answer. This reduces response length by 28\% for Qwen3-8B with only a 1.6-point accuracy drop and 40\% for Qwen3-32B with a 2.5-point drop.
    \item We compare our approach with state-of-the-art efficient reasoning methods and demonstrate consistent improvements using the OAA curve, a unified metric that accounts for overthinking.
    \item We analyze the trade-off between response length and accuracy, and study how the trained models adapt their chains of thought (CoT) for problems of varying difficulty.
\end{itemize}

\section{Methodology}
To obtain optimal LLM reasoning traces, we propose a multi-stage training framework based on: supervised fine-tuning followed by reinforcement learning with an adaptive length penalty. This approach follows the paradigm originally used to train reasoning LLMs~\citep{guo2025deepseek}.

\paragraph{Supervised Fine-Tuning} 
This first stage serves as a warm-up for RL training that also improves its convergence. We construct our supervised training datasets using the following approaches: \\
\begin{enumerate}[leftmargin=*,nosep]
    \item \textbf{Rejection sampling}: For each problem, we generate multiple continuations and select the shortest correct one. While rejection sampling has been explored in prior work as a baseline or stand-alone method \citep{yang2025towards}, in contrast, we use it as the initial stage to bias the model toward concise reasoning traces.

    \item \textbf{Reformatting}: This approach modifies the format of model-generated reasoning traces. Reasoning models typically produce a structured trace in which the intermediate reasoning (often enclosed within <think></think>) is followed by a summary and then the final answer. We construct the dataset by removing the summary and retaining only the final answer, encouraging the model to generate direct solutions without redundant reformulations.
\end{enumerate}

\paragraph{RL with Adaptive Length Penalty} After SFT, we further improve efficiency through RL with an adaptive length penalty. Specifically, we design a verifiable reward function and use group relative policy optimization (GRPO;~\citep{shao2024deepseekmath}) for training. In addition to the standard correctness reward, we apply a length penalty to encourage shorter, input-dependent reasoning traces, penalizing tokens generated after the first correct answer. Prior methods truncate or prune traces at the token or sentence level~\citep{cui2025stepwise, xia2025tokenskip}, which can disrupt the reasoning flow. In contrast, our reward function promotes responses that are concise, complete, and correct.  

The penalty is defined as the proportion of \emph{tokens after the first correct answer} relative to the full trace. Formally, let $y$ denote the generated token sequence, $y_{\text{first}}$ the subsequence up to the first correct answer (empty if none is produced), and $L$ a function returning the number of tokens in a sequence. The length penalty is:

\[
R_L(y) =
\begin{cases}
\frac{L(y) - L(y_{\text{first}})}{L(y)} & \text{if the answer is correct}, \\
0 & \text{otherwise},
\end{cases}
\]

where $y_{\text{first}}$ denotes the prefix ending at the first correct answer. Let $R_C(y)$ denote the correctness and format reward. The overall reward is

\[
R(y) = R_C(y) - \lambda R_L(y),
\]

where $\lambda$ controls the trade-off between correctness and reasoning efficiency; in our experiments, we set $\lambda=1$.

We locate the first correct answer using normalized matching. If no correct answer is produced, $y_{\text{first}} = \emptyset$, yielding zero length penalty. This discourages redundant self-verification while allowing self-correction: if the model initially produces an incorrect answer but later revises it correctly, no penalty is applied.

We refer to the method using rejection sampling during SFT as \textbf{\emph{Adaptive-Answer}}, and the method using trace reformatting during SFT as \textbf{\emph{Format-Adaptive-Answer}}.

\section{Experimental Setup}
\paragraph{Models.} We use Qwen3-8B~\citep{yang2025qwen3} as the main model in our experiments. We further validate our method on Qwen3-1.7B, Qwen3-32B and DeepSeek-R1-Qwen-7B-distilled~\citep{guo2025deepseek}. Qwen3-32B was directly fine-tuned with reinforcement learning with verifiable reward, while the other models were trained via supervised fine-tuning on reasoning traces generated by a larger model. 

\paragraph{Training Dataset.} We train models on a sample of 13K problems from DeepScaleR~\citep{deepscaler2025}, a collection of math datasets with problems drawn from AIME 1983-2023, AMC, Omni-Math \citep{gao2025omnimath}, and STILL \citep{min2024imitate}. 

Although training exclusively on math datasets, we evaluate our models across diverse domains, including science QA, commonsense reasoning, code generation, and long-context tasks. Our results show that the effects of our adaptive length penalty—reducing redundant self-verification and avoiding unnecessary continuation once correctness is reached—are domain-agnostic properties of reasoning traces. We see consistent reductions in response length with minimal accuracy loss across non-math tasks outside of our training domain.

\paragraph{Evaluation.} We evaluate the models on a diverse set of datasets covering mathematics, coding, question answering, and long-context reasoning (details in Appendix \ref{appendix:dataset}): AIME 24, AIME 25 \citep{aime}, MATH-500 \citep{lightman2023let}, LiveCodeBenchv6 \cite{jain2024livecodebench}, GPQA-Diamond \citep{rein2024gpqa}, LongBenchv2 \cite{bai2024longbench2}, and CommonsenseQA \cite{talmor2018commonsenseqa}. We use the following decoding hyperparameters as recommended in \citet{yang2025qwen3} for the Qwen3 models: temperature = $0.7$, top-p = $0.8$, top-k = $20$, and presence penalty = $1.5$. The maximum number of output tokens is set to 32,768, except for MATH-500, AIME 24, and AIME 25, where it is set to 40,000. For each question, we sample $N$ times and report the average accuracy as the final score, using $N = 64$ for AIME 24 and AIME 25, and $N = 10$ for the remaining datasets.

\paragraph{Implementation Details.} For rejection sampling, we generate 8 continuations for each problem. During the SFT stage, we train for 2 epochs with a batch size of 256 and a learning rate of 1e-5. Regarding the RL stage, we use GRPO \citep{shao2024deepseekmath} as implemented by the Verl framework \citep{sheng2024hybridflow}. We fine-tune the models with a group size of 8 and a global batch size of 256 for 50 iterations. We use the Adam optimizer with a learning rate of 1e-6, KL regularization with $\beta = 0.001$. For all experiments, including the baselines, we set the maximum number of output tokens to 16,384.

\begin{figure*}[!th]
    \centering
    \includegraphics[width=0.99\textwidth]{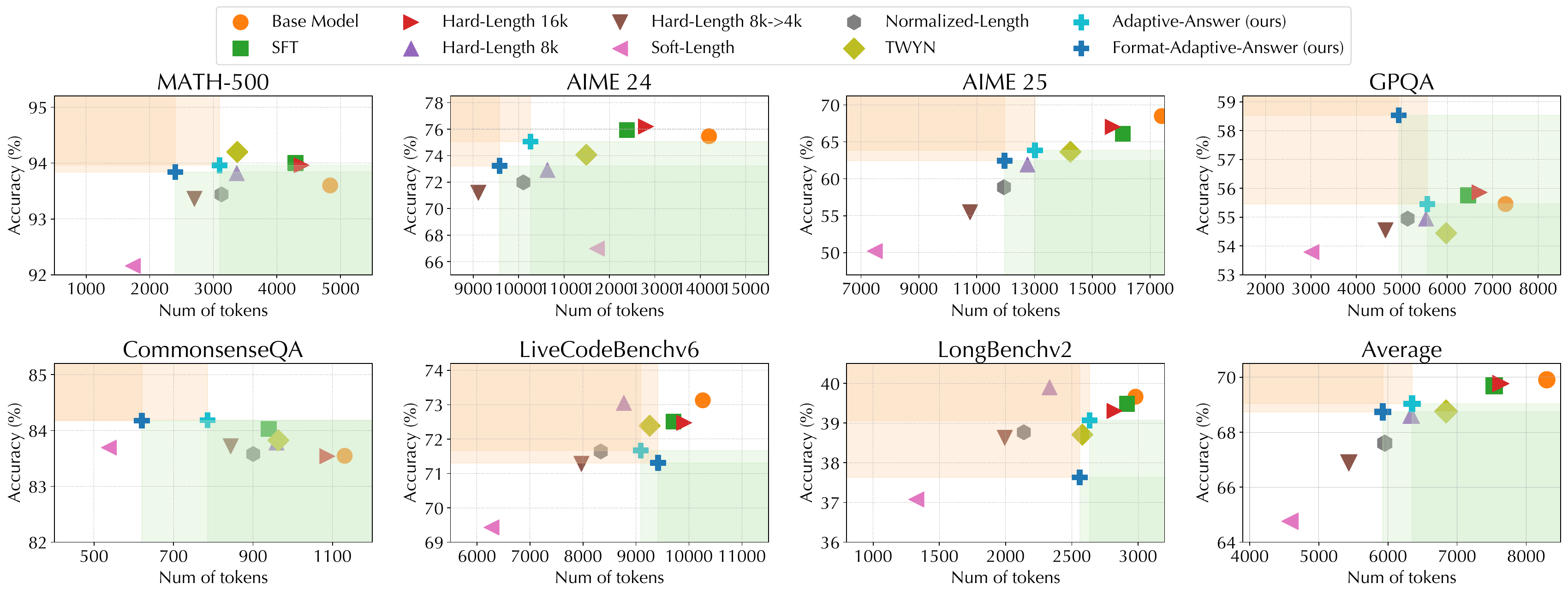}
    \caption{Average accuracy versus number of tokens for each method using Qwen3-8B. Points in the \colorbox{green!10}{green region} are dominated by \textit{Adaptive-Answer} or \textit{Format-Adaptive-Answer}, while points in the \colorbox{orange!10}{orange region} dominate them (higher accuracy, fewer tokens).}
    \label{fig:trade_off_curve}
\end{figure*}

\paragraph{Metrics.} We report both accuracy and response length (number of generated tokens) to characterize the performance-efficiency trade-off. Not all points can be directly compared using these two metrics. Therefore, we also report the area under the Overthinking-Adjusted Accuracy curve ($\text{AUC}_{\text{OAA}}$;~\citet{aggarwal2025optimalthinkingbench}). $\text{OAA}_{t}$ measures the accuracy of the model when using fewer than $t$ tokens:
\[
\text{OAA}_t = \frac{1}{n} \sum_{i=1}^n (\text{Accuracy}_i \cdot \mathbb{I} [º\text{\#Tokens}_i < t])
\]

$\text{AUC}_{\text{OAA}}$ is the area under the $\text{OAA}_{t}$ curve, where the x-axis represents the token threshold $t$ and the y-axis represents the corresponding $\text{OAA}_{t}$ score, as illustrated in Figure \ref{fig:auc}. 
\[
\text{AUC}_{\text{OAA}} = \int_0^{t_{\text{max}}} \frac{\text{OAA}_t} {t_\text{max}} dt \approx  \sum_0^{t_\text{max}} \frac{\text{OAA}_t} {t_\text{max}}
\]
where $t_\text{max}$ is a predefined maximum number of tokens. Setting $t_\text{max}$ to a very large value is equivalent to using regular accuracy, which does not account for shorter traces. Therefore, for each dataset, we set $t_\text{max}$ to the mean number of tokens generated by the original base model.

\paragraph{Baselines.} We compare our methods with existing state-of-the-art efficient reasoning approaches. We select a representative set of methods to cover a broad range of techniques:
\begin{itemize}[leftmargin=*,nosep]
    \item \textbf{No Thinking}: We disable thinking following the original Qwen3 paper \citep{yang2025qwen3}.
    
    \item \textbf{Supervised Fine-tuning (SFT)}: For each problem in the training dataset, we generate 8 continuations and retain the shortest correct answer. We then fine-tune the model on the resulting dataset. 

    \item \textbf{RL with Hard Length Penalty}~\citep{hou2025thinkprune}: Traces are truncated if they exceed a pre-defined maximum length. We set this threshold to 16k tokens, the maximum used in all our experiments, and 8k tokens, the average response length on the training set. We also report a curriculum variant that first trains with an 8k cutoff before lowering the threshold to 4k.

    \item \textbf{RL with Soft Length Penalty}~\citep{yu2025dapo}: In addition to a hard cutoff $L_\text{max}$, a second threshold $L_\text{cache}$ introduces a gradually increasing penalty once the response length exceeds it. We set $L_\text{max} = 10$k and $L_\text{cache} = 8$k.  

    \item \textbf{RL with Normalized Length Penalty}~\citep{team2025kimi}: The length penalty is normalized using the minimum and maximum response lengths sampled within the same GRPO group.  

    \item \textbf{RL with TWYN}~\citep{yang2025think}: Think When You Need (TWYN) is an adaptive method where rewards are based on pairwise comparisons: shorter correct responses receive higher rewards, while all incorrect responses receive equally low rewards. 

\end{itemize}

\section{Experimental Results}

\paragraph{Response Length Reduction.} Figure~\ref{fig:trade_off_curve} shows accuracy as a function of response length across datasets when applying different efficient reasoning methods to Qwen3-8B (see Table~\ref{tab:qwen_acc_len} in Appendix~\ref{appendix:tradeoff} for absolute values). The green region indicates points dominated by \textit{Adaptive-Answer}, while the orange region indicates points that dominate \textit{Adaptive-Answer}.

We can see that our methods substantially reduce response length while maintaining accuracy on most datasets. The degree of reduction varies across tasks; however, even the less aggressive length-reduction variant, \emph{Adaptive-Answer}, dominates other alternatives (i.e., there are almost no points in the orange area in the figures). More precisely, \emph{Adaptive-Answer} dominates most methods on MATH‑500, AIME 24, and CommonsenseQA, and is only dominated in two cases: (\emph{i})~by \textit{Hard-Length} and \textit{Soft-Length} on LiveCodeBench, and (\emph{ii})~by \textit{Hard-Length} on LongBenchv2. \emph{Format-Adaptive-Answer} dominates almost all other methods on the math and QA datasets, but is dominated on LiveCodeBench and LongBenchv2. We attribute the smaller reduction in response length across all efficient reasoning approaches on these two datasets to training primarily on math datasets.

\begin{table*}[t]
\centering
\setlength{\tabcolsep}{2pt}
\small
\resizebox{\textwidth}{!}{%
\begin{tabular}{lcccccccc}
\toprule
 & \textbf{MATH-500}
 & \textbf{AIME 24}
 & \textbf{AIME 25}
 & \textbf{\begin{tabular}[c]{@{}c@{}}GPQA\\ Diamond\end{tabular}}
 & \textbf{\begin{tabular}[c]{@{}c@{}}Common-\\ sense QA\end{tabular}}
 & \textbf{\begin{tabular}[c]{@{}c@{}}LiveCode-\\ Bench\end{tabular}}
 & \textbf{\begin{tabular}[c]{@{}c@{}}Long-\\ Benchv2\end{tabular}}
 & \textbf{Avg.} \\
\midrule
Base model
 & 81.7 & 68.6 & 75.8 & \underline{69.4} & 72.1 & \textbf{94.3} & 39.4 & 71.6 \\
\midrule
No-Thinking
 & 81.3 & 21.9 & 16.0 & 40.8 & 48.0 & 41.0 & 30.0 & 39.8 \\
SFT
 & 87.9 & 73.7 & 76.9 & 68.7 & 76.8 & 93.9 & 40.2 & 74.0 \\
Hard-Length 16k
 & 83.4 & 70.7 & 76.3 & 68.9 & 72.5 & \underline{94.1} & 40.9 & 72.3 \\
Hard-Length 8k
 & 87.9 & 73.3 & 77.2 & 63.5 & 73.3 & 93.1 & 40.7 & 72.7 \\
Hard-Length 8k $\to$ 4k
 & 89.1 & 75.1 & 69.8 & 61.0 & 76.5 & 92.7 & 38.9 & 71.8 \\
Soft-Length
 & 87.9 & 72.4 & 72.9 & 62.9 & 73.3 & 93.0 & 39.4 & 71.6 \\
Normalized-Length
 & \textbf{91.9} & 77.4 & 61.0 & 58.3 & \underline{78.6} & 90.0 & 36.6 & 70.5 \\
TWYN
 & 89.6 & 74.5 & \underline{79.6} & 67.1 & 74.8 & 93.9 & 38.8 & 74.0 \\
\midrule
\emph{Adaptive-Answer} (ours)
 & 90.3 & 75.8 & \textbf{80.0} & 68.8 & \textbf{81.4} & 93.1 & 39.4 & \underline{75.5} \\
\emph{Format-Adaptive-Answer} (ours)
 & \underline{91.2} & \textbf{81.8} & \textbf{80.0} & \textbf{71.6} & \textbf{81.3} & 93.6 & 37.4 & \textbf{76.6} \\
\bottomrule
\end{tabular}%
}
\caption{$\text{AUC}_{\text{OAA}}$ of all approaches applied to Qwen3-8B across datasets. On average, \textit{Format-Adaptive-Answer} achieves the best performance, followed by \textit{Adaptive-Answer}.}
\label{tab:auc_score}
\end{table*}

In relative terms, the largest reductions---without any performance degradation---are observed on MATH-500 (36\% for \textit{Adaptive-Answer} and 50\% for \textit{Format-Adaptive-Answer}) and CommonsenseQA (30\% and 45\%, respectively). On GPQA Diamond, AIME 24, and AIME 25, response lengths decrease by about 25\% for \textit{Adaptive-Answer} and 32\% for \textit{Format-Adaptive-Answer}. For LiveCodeBench and LongBenchv2, both methods show only minor accuracy drops (less than two points) with smaller length reductions—11\% and 12\% for \textit{Adaptive-Answer}, and 8\% and 14\% for \textit{Format-Adaptive-Answer}. On average, {\textit{Adaptive-Answer} shortens responses by 28\% and \textit{Format-Adaptive-Answer} by 33\%, with only a one-point decrease in accuracy}. We must note, that even though training is performed only on math datasets, our methods also shorten responses on science, coding, QA, and long-context reasoning datasets.

We must highlight that not all methods are directly comparable using accuracy and response length. For example, on all datasets except AIME 24, \textit{Soft-Length} neither dominates nor is dominated by other methods. Therefore, we also compare the $\text{AUC}_{\text{OAA}}$ of all approaches applied to Qwen3-8B across datasets (see Table~\ref{tab:auc_score}). This confirms the effectiveness of our methods: on average, \textit{Format-Adaptive-Answer} outperforms all other methods, followed by \textit{Adaptive-Answer}. \textit{Format-Adaptive-Answer} achieves the highest score on all math and question answering datasets, except for MATH-500 where it ranks second. Interestingly, on LiveCodeBench and LongBenchv2, simple baselines such as \textit{SFT} and \textit{Hard-Length 16k}—equivalent to RL without a length penalty—outperform all other efficient reasoning alternatives.

\begin{figure*}[t]
    \centering
    \includegraphics[width=0.95\linewidth]{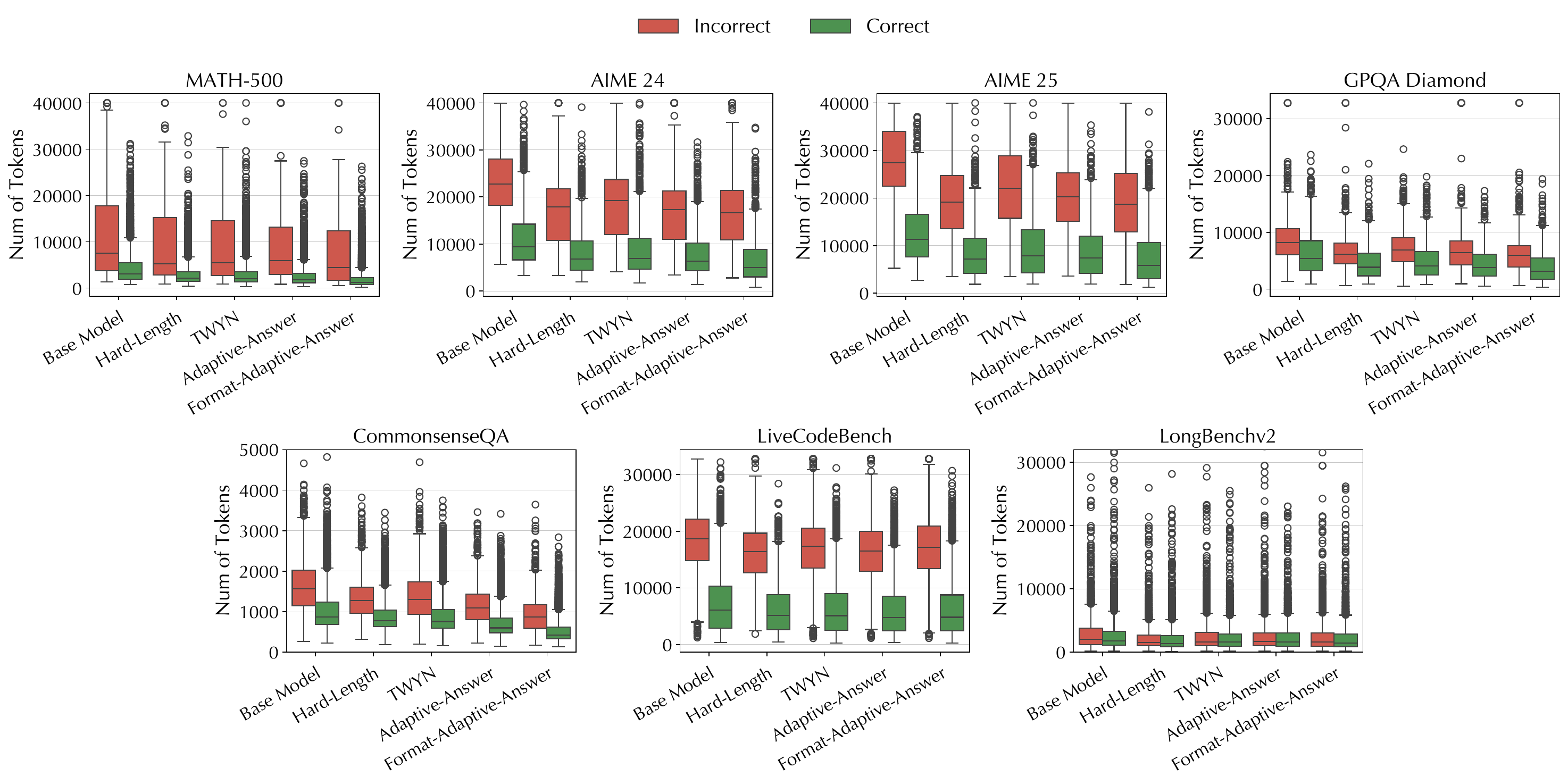}
    \caption{Response length distributions of some representative efficient reasoning methods applied to Qwen3-8B. We separate the correct and incorrect responses.}
    \label{fig:token_analysis}
\end{figure*}

\begin{table}[!t]
\begin{adjustbox}{width=1\linewidth}
\begin{tabular}{lccc}
\toprule
         & \textbf{Accuracy} & \textbf{\#Tokens} & \textbf{AUC\textsubscript{OAA}}   \\
\midrule
Base Model & 69.9  & 8295  & 71.6  \\
\midrule
SFT (Rejection Sampling) & 69.7  & 7536  & 74.0 \\
SFT (Formatting) & 70.1  & 7475  & 74.7 \\
RL (no SFT) & 68.8  & 6303  & 73.2 \\
\midrule
\emph{Adaptive-Answer} & 69.0 & 6344 & 75.5\\
\emph{Format-Adaptive-Answer} & 68.7 & 5918 & 76.6 \\
\bottomrule
\end{tabular}
\end{adjustbox}
\caption{Average accuracy, response length, and $\text{AUC}_{\text{OAA}}$ of the original model, rejection sampling–based SFT, format-based SFT, RL with adaptive length penalty (without SFT), rejection sampling–based SFT followed by RL (\textit{Adaptive-Answer}), and format-based SFT followed by RL (\textit{Format-Adaptive-Answer}).}
\label{tab:ablation}
\end{table}

\paragraph{Component Ablations.} 
To evaluate the contribution of each component of our approach, we perform ablations for each component individually. Table~\ref{tab:ablation} reports the average accuracy, response length, and $\text{AUC}_{\text{OAA}}$ for several configurations: the original model, rejection sampling–based SFT, format-based SFT, RL with adaptive length penalty (without SFT), rejection sampling–based SFT followed by RL (\textit{Adaptive-Answer}), and format-based SFT followed by RL (\textit{Format-Adaptive-Answer}).

Adding the rejection sampling–based SFT (\emph{SFT~(Rejection Sampling)}) stage does not yield clear improvements when examining accuracy or response length alone—\emph{Adaptive-Answer} achieves slightly higher accuracy than \textit{RL (no SFT)} but produces longer responses. Hence, we argue that $\text{AUC}_{\text{OAA}}$ is a more suitable metric for ranking models when no model clearly dominates another. $\text{AUC}_{\text{OAA}}$ clearly highlights the benefit of the SFT phase. We observe a sizable improvement of 3 AUC points over the base model.

Format-based SFT (\emph{SFT (Formatting)}) reduces response length by  10\% on average without loss in accuracy. This suggests that the summary generated before the final answer does not contribute to performance, as the model reaches the correct answer by the end of the trace. Adding the RL stage further improves $\text{AUC}_{\text{OAA}}$ by 3 points and cuts response length 18\%, with only a minor drop in accuracy. However, combining rejection sampling with formatting during SFT does not yield additional improvements over formatting alone.

Finally, SFT is a crucial stage for RL performance: although \emph{RL (no SFT)} produces the second-shortest reasoning traces, it incurs a performance penalty and achieves a lower $\text{AUC}_{\text{OAA}}$ compared to the full approaches (\emph{Adaptive-Answer} and \emph{Format-Adaptive-Answer}).

\begin{table}[t!]
\begin{adjustbox}{width=1\linewidth}
\begin{tabular}{lclc}
\toprule
         & \textbf{Accuracy} & \textbf{\#Tokens} & \textbf{AUC\textsubscript{OAA}}   \\
\midrule 
\multicolumn{4}{c}{\textbf{Qwen3-1.7B}} \\
\midrule
Base Model  & 50.9  & 7,619  & 65.4  \\
Adaptive-Answer & 49.1 & 5,884 (-22\%) & 62.1\\
Format-Adaptive-Answer & 48.3 & 5,918 (-22\%) & 62.1 \\
\midrule
\multicolumn{4}{c}{\textbf{Qwen3-8B}} \\
\midrule
Base Model  & 69.9  & 8,298  & 71.6  \\
Adaptive-Answer & 69.0 & 6,344 (-23\%) & 75.5\\
Format-Adaptive-Answer & 68.7 & 5,918 (-28\%) & 76.6 \\
\midrule
\multicolumn{4}{c}{\textbf{Qwen3-32B}} \\
\midrule
Base Model  & 74.8  & 7,294  & 69.2  \\
Adaptive-Answer & 72.1 & 4,280 (-41\%) & 72.5\\
Format-Adaptive-Answer & 72.2 & 4,372 (-40\%) & 72.3 \\
\midrule
\multicolumn{4}{c}{\textbf{DeepSeek-R1-Qwen-7B-Distill}} \\
\midrule
Base Model & 50.3  & 6,272  & 62.1  \\
Adaptive-Answer
 & 50.4 & 5133 (18\%) & 59.6\\
Format-Adaptive-Answer & 50.7 & 4,612 (26\%)  & 59.7 \\
\bottomrule
\end{tabular}
\end{adjustbox}
\caption{Average accuracy, response length and $\text{AUC}_{\text{OAA}}$ of our methods applied to Qwen3.1.7B, Qwen3-8B, Qwen-32B and DeepSeek-R1-Qwen-7B-distilled. }
\label{tab:auc_all}
\end{table}

\paragraph{Efficiency and Model Size.} We investigate how our approaches scale with model size and training regimes. For this set of experiments, we evaluate Qwen3-\{1.7B, 8B, 32B\}, and DeepSeek-R1-Qwen-7B-Distilled. 

Across all models, the performance drop after applying our methods remains under 2.5 points—and is negligible for DeepSeek-R1-7B (see Table~\ref{tab:auc_all}). {Notably, the reduction in generated tokens increases with model size: \textit{Adaptive-Answer} shortens responses by 22\% on Qwen3-1.7B, 23\% on Qwen3-8B, and 40\% on Qwen3-32B}, demonstrating that larger models benefit more from efficient reasoning. However, $\text{AUC}_{\text{OAA}}$ {does not always align perfectly with the accuracy–response length trade-offs, particularly for smaller models}, highlighting that efficiency gains can sometimes come at a subtle cost to overall reasoning effectiveness. For instance, the fine-tuned DeepSeek-R1 dominates the base model in absolute accuracy but achieves a slightly lower $\text{AUC}_{\text{OAA}}$ score. {Overall, these results indicate that our methods are model-agnostic, consistently effective across different model sizes, and operating without significant performance loss}.

\begin{figure*}[t]
    \centering
    \includegraphics[width=0.9\linewidth]{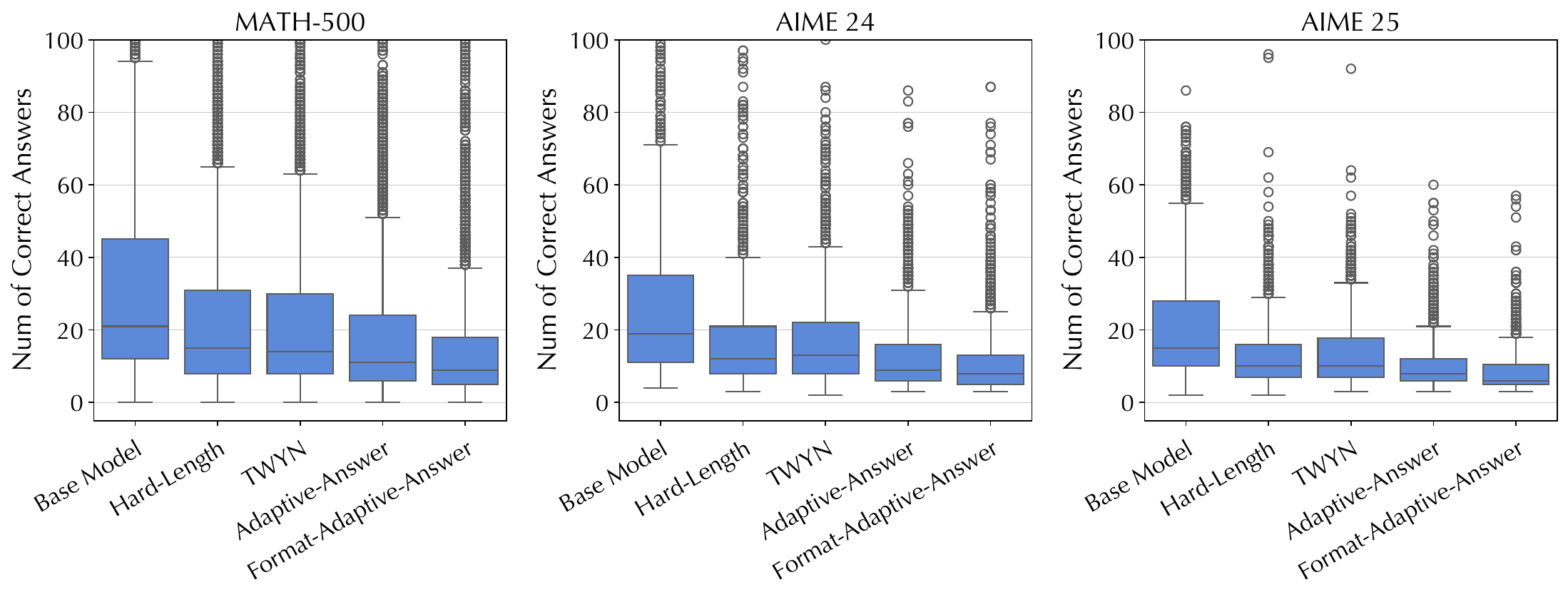}
    \caption{Distributions of the number of correct answers in the traces of some representative efficient reasoning methods applied to Qwen3-8B for MATH-500, AIME 24 and AIME 25.}
    \label{fig:answer_analysis}
\end{figure*}

\begin{figure*}[t]
    \centering
    \includegraphics[width=0.9\linewidth]{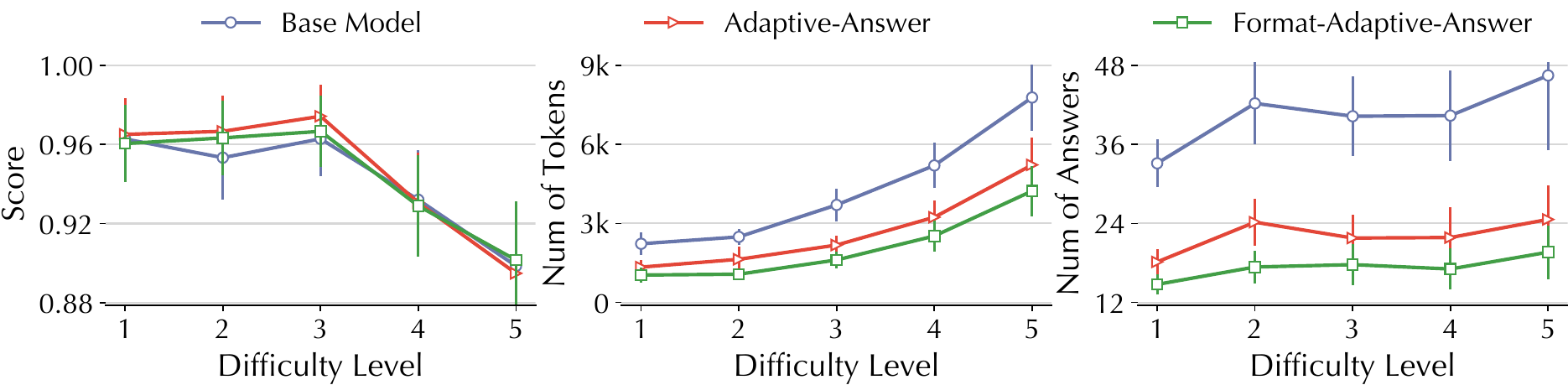}
    \caption{Accuracy, response length, and count of intermediate correct steps across difficulty levels on MATH-500.}
    \label{fig:diff_analysis}
\end{figure*}

\begin{figure*}[t]

\begin{tcolorbox}[colback=white!5!white,colframe=black!75!black]
    {\scriptsize
    \begin{tcolorbox}[title=Base Model]
    \hlblue{<think>Okay, so Jen is playing this lottery game ... So yes, $m=1$ and $n=115$, so $m+n = \boxed{\textbf{116}}$.}
    \hlgreen{But wait, that seems low ... Then $m+n = 1+115 = \boxed{\textbf{116}}$.}
    \hlyellow{Hmm. But I feel like this might not be correct. … Therefore, $m+n = 1+115 = \boxed{\textbf{116}}$.}
    \hlred{But I need to make sure that this is correct. Let me think again … Therefore, $m+n = 1+115 = \boxed{\textbf{116}}$.}
    \hlcyan{But wait, intuitively, if the probability of winning the grand prize given …  
    Therefore, the answer is $\boxed{\textbf{116}}$.}
    \hlmagenta{But wait, let me check again … 
    Therefore, answer is $1+115 = \boxed{\textbf{116}}$.}
    \hlorange{But let me check if there's an alternative approach... so the answer is $1/115$, so $m+n = 1+115 = \boxed{\textbf{116}}$.}
    \hlblue{But just to be thorough, let me check if …   Final Answer: $\boxed{\textbf{116}}$<think>} 
    \hlgreen{To solve this problem, we are asked to find the **conditional probability** … and their sum is:$m + n = 1 + 115 = \boxed{\textbf{116}}$}
    \end{tcolorbox}

    \begin{tcolorbox}[title=Adaptive-Answer]
        \hlblue{<think>Okay, so Jen is entering a lottery where she has to pick 4 distinct numbers from the set S … So m + n is 1 + 115 = $\boxed{\textbf{116}}$.}
        \hlgreen{Wait, but before I conclude, let me verify once more because sometimes with probability … Therefore, the answer is 1 + 115 = $\boxed{\textbf{116}}$.}
        \hlyellow{But just to make sure, let me think of another way … I think the answer is 116.**Final Answer**$\boxed{\textbf{116}}$</think>}
        \hlred{To solve the problem, we are asked to find the probability … These are relatively prime.$m + n = 1 + 115 = \boxed{\textbf{116}}$}
    \end{tcolorbox}

    \begin{tcolorbox}[title=Format-Adaptive-Answer]
        \hlblue{<think>Okay, let me try to figure out this probability problem. So, Jen is entering a lottery where … **Final Answer**$\boxed{\textbf{116}}$</think>}
        \hlgreen{**Final Answer**$\boxed{\textbf{116}}$.}
    \end{tcolorbox}
    }
\end{tcolorbox}
\caption{Reasoning traces of Qwen3-8B on AIME 24 Problem 10 before and after fine-tuning. The base model performs seven self-verifications after arriving at the correct answer, whereas \textit{Adaptive-Answer}  performs only two and \textit{Format-Adaptive-Answer}  performs none.}
\label{fig:example_aime24}
\end{figure*}

\section{Analysis}

\paragraph{Response Length Distribution.} Figure~\ref{fig:token_analysis} shows the response length distributions for a representative set of efficient reasoning methods applied to Qwen3-8B. {Across all three datasets, incorrect answers tend to have longer traces than correct ones}, highlighting a correlation between excessive reasoning and errors. {Importantly, our methods effectively shift the response length distribution for both correct and incorrect answers}, showing that the models adapt traces consistently, regardless of the final answer. {This indicates that our approach encourages concise reasoning across all outputs, not just the correct ones}.

\paragraph{Intermediate Answers.} 
The RL stage encourages the model to minimize unnecessary self-verification. To analyze this, we report the number of correct answers appearing in each reasoning trace for MATH-500, AIME 24, and AIME 25 (Figure~\ref{fig:answer_analysis}). While this metric is a coarse proxy—since answers may be repeated or paraphrased—it provides qualitative insight into verification behavior. Both \emph{Adaptive-Answer} and \emph{Format-Adaptive-Answer} {shift the distribution toward fewer intermediate correct answers}, indicating reduced redundancy in reasoning.

\paragraph{Difficulty Analysis.} 
We examine how problem difficulty affects accuracy, response length, and intermediate correct answers. Each MATH-500 problem is assigned a difficulty level from 1 to 5. As shown in Figure~\ref{fig:diff_analysis}, our approaches maintain the base model’s accuracy across all levels. Response length adapts to difficulty, increasing for harder problems, reflecting the need for more reasoning. Although both response length and intermediate correct answers rise with difficulty, {they remain shorter than the base model}, demonstrating our methods’ efficiency even on challenging problems.

\paragraph{Qualitative Analysis.} 
Figure~\ref{fig:example_aime24} compares reasoning traces for a math problem from AIME24 produced by: (\emph{i})~the base model, (\emph{ii})~\emph{Adaptive-Answer}, and (\emph{iii}) \emph{Format-Adaptive-Answer} (long responses are trimmed). We see that the base model performs seven unnecessary self-verifications after producing the first correct answer (\emph{116}). In contrast, \emph{Adaptive-Answer} reduces this to three, while \emph{Format-Adaptive-Answer} produces an optimal trace with no self-verifications, directly generating the final answer without summarizing the reasoning.

\section{Related Work}
\paragraph{Test-Time Scaling.} Large Language Models perform better on reasoning-heavy tasks such as math, problem-solving, and coding by increasing test-time computation \citep{wei2022chain, wang2022self, snell2025scaling}. Models generate intermediate tokens in parallel—by sampling multiple traces \citep{wang2022self}—or sequentially, by verifying and correcting their own outputs \citep{madaan2023self, kumar2024training}. Recent works use reinforcement learning with verifiable rewards to further enhance reasoning capabilities, leading to longer CoT as well as self-verification and self-correction behaviors \citep{jaech2024openai, guo2025deepseek, openai2025, yang2025qwen3}. 

\paragraph{Efficient Reasoning.} While reinforcement learning improves reasoning ability, it often comes at the cost of efficiency. In some cases, reasoning traces become excessively long, increasing computation without improving accuracy—and sometimes even harming it, a phenomenon known as ``overthinking'' \citep{chen2025do, yang2025towards, wu2025more}. Several methods have been proposed to address this issue.
The most direct approach, budget forcing \citep{muennighoff2025s1, yang2025qwen3}, interrupts generation once a predefined threshold is exceeded. Other methods \citep{yang2025towards, xia2025tokenskip, cui2025stepwise} construct synthetic datasets by shortening model-generated reasoning traces (via rejection sampling or pruning) and then perform supervised fine-tuning.
A different line of work, to which our method belongs, uses reinforcement learning with a length penalty in addition to the correctness reward \citep{lou2025adacot, zhang2025adaptthink}. The length constraint can be either hard, applied once the CoT length exceeds a fixed threshold \citep{hou2025thinkprune}, or soft, where the penalty increases gradually as the trace length approaches the threshold \citep{yu2025dapo, aggarwal2025l1}. Instead of applying penalties independently per example, some approaches \citep{team2025kimi, yang2025think} define them relative to the length and correctness of other traces within the same GRPO group.

Unlike prior methods, which rely on a manually fixed ``thinking budget'' shared across all inputs, we train models to produce short yet complete reasoning traces while preserving accuracy. Our approach incentivizes the model to adaptively infer an input-dependent budget. %

\section{Conclusion}
Large Language Models (LLMs) often perform better on reasoning-intensive tasks by producing longer chains of thought. However, these chains are often unnecessarily long, increasing inference costs without improving accuracy. To address this, we propose a multi-stage efficient reasoning framework that consists of supervised fine-tuning—via rejection sampling or reformatting—followed by reinforcement learning with an adaptive length penalty. Our approach effectively shortens response length (28\% for Qwen3-8B and 40\% for Qwen3-32B) with only minor performance drops (up to 2.5 points accuracy) and outperforms existing state-of-the-art efficient reasoning methods by 2.5 points when evaluated with the unified metric $\text{AUC}_{\text{OAA}}$.

\section*{Limitations}
Although we evaluate our methods on datasets from multiple domains, our training is performed exclusively on math datasets. Extending training to a more diverse set of tasks could yield a better accuracy–response length trade-off. Moreover, the efficient reasoning methods we propose are post hoc interventions; we do not explore incorporating the adaptive length penalty directly during the initial RL training. Additionally, due to resourcing constraints, we focused our experiment scope to models of different sizes within the Qwen family based on a dense architecture. Future explorations can consider extending our approach to more model families and other architectures such as Mixture-of-Experts.  Finally, we focus exclusively on the model performance on reasoning tasks, and we do not measure the change in performance on other task groups that are less dependent on the CoT quality.

\section*{Ethics Statement}
One of our methods removes the summary that the model produces at the end of its thinking content. Although this results in shorter responses, this might also reduce the legibility of the reasoning traces.

\section*{Acknowledgments}
UPF was funded by the European Research Council (ERC) under the European Union’s Horizon 2020 research and innovation programme (grant agreement No. 101019291). This paper reflects the authors’ view only, and the funding agency is not responsible for any use that may be made of the information it contains.

\bibliography{custom}

\clearpage
\appendix
\section{Datasets}
\label{appendix:dataset}
We evaluate on the following datasets that come from diverse domains including math, science, coding, question answering and long context reasoning:
\begin{itemize}[leftmargin=0pt, label={}]
    \item \textbf{MATH-500} \citep{lightman2023let}: A representative subset of 500 problems from the MATH benchmark. Each problem is assigned a difficulty level ranging from 1 to 5. 
    \item \textbf{AIME 24} \citep{aime}: 30 math problems from the 2024 edition of the American Invitational Mathematics Examination, a prestigious high school mathematics competition known for its challenging mathematical problems.
    \item \textbf{AIME 25} \citep{aime}: 30 math problems from the 2025 edition of the American Invitational Mathematics Examination, a prestigious high school mathematics competition known for its challenging mathematical problems.
    \item \textbf{GPQA Diamond} \citep{rein2024gpqa}: a subset of 198 expert-written, graduate-level questions in biology, physics, and chemistry, designed to test the true reasoning abilities of Large Language Models (LLMs) without reliance on easily found internet answers
    \item \textbf{CommonsenseQA} \citep{talmor2018commonsenseqa}: a dataset consisting of 1221 multiple-choice questions that require commonsense knowledge to predict the correct answers . Each question has one correct answer and four distractor answers.
    \item \textbf{LiveCodeBench} \citep{jain2024livecodebench}: A holistic and contamination-free benchmark to evaluate the coding capabilities of LLMs. We use the sixth version of the dataset which contains 055 problems. 
    \item \textbf{LongBenchv2} \citep{bai2024longbench2}: a dataset of 503 challenging multiple-choice questions, with contexts ranging from 8k to 2M words. It contains the following categories: single-document QA, multi-document QA, long in-context learning, long-dialogue history understanding, code repo understanding, and long structured data understanding. 
\end{itemize}

\section{Accuracy-Response Length Trade-off}
\label{appendix:tradeoff}

Table~\ref{tab:qwen_acc_len} quantifies the accuracy--length trade-off. Tight hard-length constraints reduce average response length from 8.3k tokens (Base Model) to 5.4k, but incur a 3.0-point average accuracy drop and a severe degradation on AIME~25 (68.5 $\rightarrow$ 55.5). Normalized-Length achieves the shortest outputs (4.6k tokens on average) but suffers the largest performance loss (69.9 $\rightarrow$ 64.8). In contrast, adaptive methods preserve accuracy more effectively: Adaptive-Answer reduces average length by 23.5\% (8.3k $\rightarrow$ 6.3k) with only a 0.9-point accuracy decrease, while Format-Adaptive-Answer achieves a 28.6\% reduction (5.9k tokens) with a 1.2-point drop. Among all efficient reasoning strategies, our proposed methods consistently occupy the Pareto-optimal region, yielding the best overall accuracy--efficiency trade-offs across benchmarks. These results indicate that instance-level length adaptation yields a substantially better efficiency--accuracy trade-off than fixed or normalized constraints.

\begin{table*}[t]
\centering
\small
\setlength{\tabcolsep}{3pt}
\resizebox{\textwidth}{!}{%
\begin{tabular}{l|rr|rr|rr|rr|rr|rr|rr|rr}
\toprule
\textbf{Model} &
\multicolumn{2}{c|}{\textbf{MATH-500}} &
\multicolumn{2}{c|}{\textbf{AIME 24}} &
\multicolumn{2}{c|}{\textbf{AIME 25}} &
\multicolumn{2}{c|}{\textbf{\begin{tabular}[c]{@{}c@{}}GPQA\\ Diamond\end{tabular}}} &
\multicolumn{2}{c|}{\textbf{\begin{tabular}[c]{@{}c@{}}Common-\\ senseQA\end{tabular}}} &
\multicolumn{2}{c|}{\textbf{\begin{tabular}[c]{@{}c@{}}LiveCode-\\ Bench\end{tabular}}} &
\multicolumn{2}{c|}{\textbf{\begin{tabular}[c]{@{}c@{}}Long-\\ Benchv2\end{tabular}}} &
\multicolumn{2}{c}{\textbf{Average}} \\
\cmidrule(lr){2-17}
& Acc.$\uparrow$ & \#Tok.$\downarrow$
& Acc.$\uparrow$ & \#Tok.$\downarrow$
& Acc.$\uparrow$ & \#Tok.$\downarrow$
& Acc.$\uparrow$ & \#Tok.$\downarrow$
& Acc.$\uparrow$ & \#Tok.$\downarrow$
& Acc.$\uparrow$ & \#Tok.$\downarrow$
& Acc.$\uparrow$ & \#Tok.$\downarrow$
& Acc.$\uparrow$ & \#Tok.$\downarrow$ \\
\midrule
Base Model
& 93.6 & 4{,}837
& 75.5 & 14{,}191
& \textbf{68.5} & 17{,}402
& 55.5 & 7{,}284
& 83.5 & 1{,}130
& \textbf{73.1} & 10{,}261
& 39.7 & 2{,}980
& \textbf{69.9} & 8{,}298 \\
\midrule
SFT
& 94.0 & 4{,}292
& 75.9 & 12{,}381
& 66.1 & 16{,}058
& 55.8 & 6{,}459
& 84.0 & 939
& 72.5 & 9{,}708
& 39.5 & 2{,}915
& 69.7 & 7{,}536 \\
Hard-Length 16k
& 94.0 & 4{,}388
& \textbf{76.2} & 12{,}803
& 67.0 & 15{,}707
& 55.9 & 6{,}712
& 83.5 & 1{,}087
& 72.5 & 9{,}909
& 39.3 & 2{,}822
& 69.8 & 7{,}633 \\
Hard-Length 8k
& 93.8 & 3{,}369
& 72.9 & 10{,}633
& 61.9 & 12{,}758
& 54.9 & 5{,}535
& 83.8 & 959
& \textbf{73.1} & 8{,}772
& \textbf{39.9} & 2{,}332
& 68.6 & 6{,}337 \\
Hard-Length 8k $\rightarrow$ 4k
& 93.4 & 2{,}703
& 71.2 & \textbf{9{,}114}
& 55.5 & 10{,}770
& 54.5 & 4{,}642
& 83.7 & 844
& 71.3 & 7{,}972
& 38.6 & 1{,}994
& 66.9 & 5{,}434 \\
Soft-Length
& 93.4 & 3{,}129
& 72.0 & 10{,}105
& 58.9 & 11{,}944
& 54.9 & 5{,}128
& 83.6 & 900
& 71.6 & 8{,}335
& 38.8 & 2{,}137
& 67.6 & 5{,}954 \\
Normalized-Length
& 92.2 & \textbf{1{,}734}
& 67.0 & 11{,}723
& 50.2 & \textbf{7{,}475}
& 53.8 & \textbf{3{,}011}
& 83.7 & \textbf{537}
& 69.4 & \textbf{6{,}273}
& 37.1 & \textbf{1{,}326}
& 64.8 & \textbf{4{,}583} \\
TWYN
& \textbf{94.2} & 3{,}377
& 74.1 & 11{,}491
& 63.6 & 14{,}243
& 54.4 & 5{,}978
& 83.8 & 964
& 72.4 & 9{,}259
& 38.7 & 2{,}579
& 68.8 & 6{,}841 \\
\midrule
\textbf{Adaptive-Answer}
& 94.0 & 3{,}098
& 75.1 & 10{,}261
& 63.9 & 13{,}017
& 55.5 & 5{,}560
& \textbf{84.2} & 786
& 71.7 & 9{,}089
& 39.1 & 2{,}634
& 69.0 & 6{,}349 \\
\textbf{Format-Adaptive-Answer}
& 93.8 & 2{,}403
& 73.2 & 9{,}583
& 62.4 & 11{,}965
& \textbf{58.5} & 4{,}931
& \textbf{84.2} & 620
& 71.3 & 9{,}416
& 37.6 & 2{,}559
& 68.7 & 5{,}925 \\
\bottomrule
\end{tabular}
}
\caption{Accuracy (Acc.$\uparrow$) and response length (\#Tok.$\downarrow$) of all approaches applied to Qwen3-8B. Best values per column are in bold.}
\label{tab:qwen_acc_len}
\end{table*}

\paragraph{AI use disclosure:} we used AI for assistance in code writing and in manuscript typesetting.

\end{document}